\PassOptionsToPackage{dvipsnames}{xcolor}
\documentclass[11pt]{article}

\usepackage[preprint]{acl}

\usepackage{times}
\usepackage{latexsym}

\usepackage[T1]{fontenc}
\usepackage[utf8]{inputenc}

\usepackage{microtype}

\usepackage{inconsolata}

\usepackage{graphicx}
\usepackage{amsmath,amssymb,amsthm}
\usepackage{amsfonts}
\usepackage{nicefrac}
\usepackage{booktabs}
\usepackage{xcolor}
\usepackage[most]{tcolorbox}
\usepackage{multirow}
\usepackage{xspace}
\usepackage{subcaption}
\usepackage{CJKutf8}

\newcommand{\Cop}{\ensuremath{\mathtt{C}}\xspace}
\newcommand{\Def}{\ensuremath{\mathtt{D}}\xspace}
\newcommand{\thinktag}{\texttt{<think>}\xspace}
\def\endthinktag{\texttt{</think>}\xspace}

\newcounter{promptboxctr}
\newtcolorbox{promptbox}[1]{
  enhanced, breakable, colback=gray!4, colframe=gray!55!black,
  boxrule=0.4pt, arc=1pt, left=6pt, right=6pt, top=4pt, bottom=4pt,
  fonttitle=\bfseries\small, title={Box~\thepromptboxctr: #1},
  fontupper=\ttfamily\small
}

\title{Tracking Representations in LLM Reasoning Traces for Iterated Games}

\author{Chaipat Tirapongprasert \\
  Columbia University \\
  \texttt{t.chaipat@columbia.edu} \\
  \And
  Rajarshi Das \\
  MQube Cognition \\
  \texttt{rajarshi.das@mqube.ai} \\
  \And
  Parijat Dube \\
  Columbia University \\
  \texttt{pd2637@columbia.edu} \\}

\begin{document}
\maketitle
\begin{abstract}
Large language model (LLM) agents achieve state-of-the-art performance across domains through chain-of-thought (CoT) reasoning. Despite their ubiquitous use, little is understood about the relationship between the externalized trace and the model's actual computation. We address this by investigating how LLM agents make decisions in iterative games with incomplete information. Using a series of mechanistic interpretability techniques, we find that the model forms and iteratively revises an internal representation of the game throughout the CoT.
Once the model prematurely commits to an incorrect belief, it rarely recovers. This failure is manifested in both the literal trace and the layer activations. Ultimately, as multi-agent systems are more prevalent, our work opens the door to a more focused diagnosis of their strategic reasoning errors. For illustrative purposes, we offer an interactive visualizer of our methodology on selected sample traces.\footnote{\url{https://interp-self.github.io/gametrace/}}
\end{abstract}

\section{Introduction}

Large language models (LLMs) of today~\citep{openai2024o1,deepseekai2025r1,anthropic2025claude37,yang2025qwen3} have shown remarkable abilities as autonomous agents capable of making complex, high-stakes decisions in a wide range of domains~\citep{yang2025codetothink,hakimov2025price,shi2026pandemic}. To achieve this state-of-the-art performance, models are trained to externalize their thinking and generate immediate reasoning steps as a long, serial chain of thought (CoT) before finalizing the answer~\citep{wei2022chain}. Despite widespread adoption, our understanding of the agents' thought processes~\citep{kosinski2023evaluating,ullman2023large,das2024iterative,das2025representationalignment} and the possible ways they can fail or be exploited~\citep{abdelnabi2024cooperation,duan2024gtbench} is still limited.

Prior work has used game theory as a tool to investigate how an LLM agent thinks and behaves under uncertainty in repeated games in which it must compete or cooperate to achieve its objectives~\citep{akata2023playing,fontana2024nicer,guo2023gpt,willis2025cooperate}. However, these traditional methods have primarily focused on the LLM's external behavior (i.e., the final outcome of a given round), which is insufficient to comprehend the strengths and/or weaknesses of its strategic decision-making. We advance this line of research by surfacing the internal mechanisms that inform the model's decision-making and identifying areas where they deviate from game-theoretic rationality.

We present a three-tiered interpretability pipeline for analyzing intermediate belief formation along the reasoning trajectory. \textbf{First,} we decompose each round's CoT into sentences and at each chunk, use multiple-choice probes to elicit a verbalized readout of belief and action. A common concern with this type of CoT interpretability work~\citep{lanham2023measuring,bogdan2025thought} is that the LLM-generated trace might not represent the genuine reasoning process that underpins the model's predictions~\citep{kambhampati2026positionstopanthropomorphizingintermediate,turpin2023language,arcuschin2025chainofthought,chen2025reasoning}. Taking this criticism seriously, we choose to unembed the residual stream at the final layer to obtain the next-token logit distribution at each chunk boundary rather than reading off the literal output text. \textbf{Second,} we study the causal dependencies between sentences by removing each individual sentence in the CoT and measuring the resulting shift in downstream logits. This step allows us to filter away performative sentences that are not mechanistically relevant to the output.
\textbf{Third,} we look into the model's intermediate activations to decode hidden representations that the LLMs might leverage during deliberation but never explicitly regurgitate in the reasoning trace.

\begin{figure}
    \centering
    \includegraphics[width=\columnwidth]{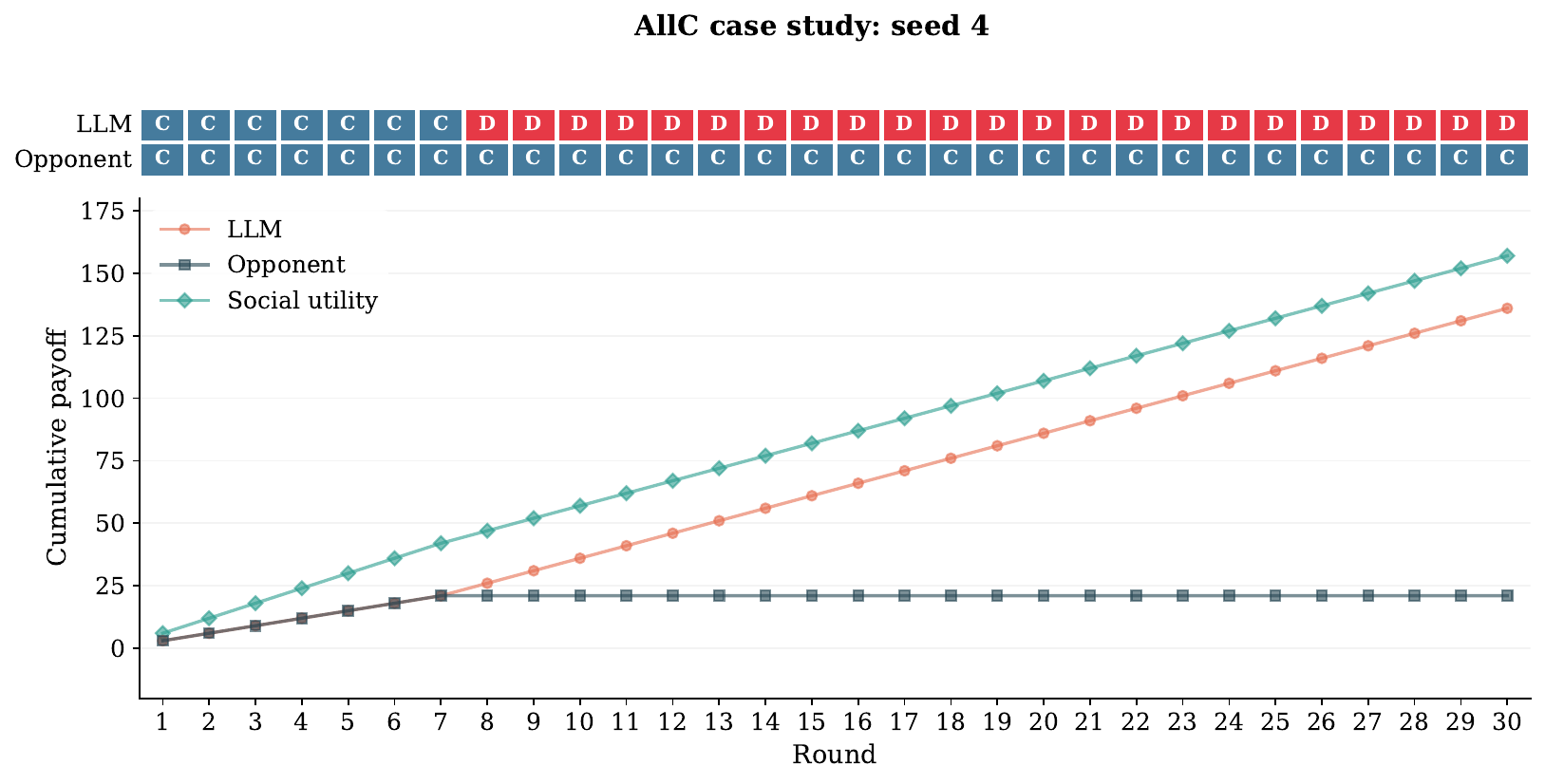}
    \caption{A full gameplay against $\textsc{AllC}$. Top: per-round decisions for the LLM and its opponent (blue = Cooperate, red = Defect). Bottom: cumulative payoff over the episode for the LLM, the opponent, and social utility (their sum). During the first 7 rounds, the LLM plays \Cop, and both players' cumulative payoffs increase along the mutual-cooperation trajectory. Starting at round 8, the LLM switches to \Def (the best strategy for maximizing the individual's payoff) and never returns to cooperation for the next 23 rounds.}
    \label{fig:gameplay}
\end{figure}

Applying these techniques, our research suggests that LLM agents develop some internal beliefs about the state of the world, which are then used to inform their behavior. Such representations are maintained and updated throughout the CoT as new information emerges. Although models may omit some incriminating reasoning in the CoT, such as fear of exploitation or punishment, these unspoken concepts are present throughout the activations of the early and middle layers.

Because the model's final decision is tightly coupled with these accumulated beliefs (verbalized or not), the belief trajectory within a trace tends to lead the action. Thus, reaching the best-response action generally entails maintaining the correct belief. We also find that the LLM's irrational behaviors are often associated with its premature commitment to a hypothesis, which is a self-reinforcing process that worsens over time. Together, we see our work as paving the way for more granular debugging of misbehavior and reasoning failures, which will become increasingly important as LLMs are deployed in complex settings with real-world consequences.

\section{Experimental Setup}
\label{sec:setup}

\paragraph{Task.} We choose to study the LLM's behavior in a two-player iterated Prisoner's Dilemma against a fixed opponent.
In this game, both players choose whether to defect (\Def) or cooperate (\Cop) at each round and receive payoffs based on their joint outcome.
The game is a dilemma in that the payoff matrix $T > R > P > S$ (see Appendix~\ref{app:ipd}) incentivizes each rational, self-interested player to defect, despite both parties winding up worse off than if they had cooperated~\citep{rapoport1965prisoner,axelrod1981evolution}.
However, if the player is patient enough and prioritizes future utility, mutual cooperation can emerge as the optimal strategy in a game that is played indefinitely (or, more practically, for an unknown number of times) ~\citep{friedman1971supergames,fudenberg1986folk}. Unfortunately, such an equilibrium relies on trust, which for both humans~\citep{dalbo2005shadow,dalbo2011evolution,axelrod1981evolution} and LLMs~\citep{akata2023playing,fontana2024nicer,guo2023gpt}, is difficult to restore after the opponent's betrayal.

\paragraph{Model and game protocol.}

We use \texttt{DeepSeek-R1-Distill-Qwen-14B}~\citep{deepseekai2025r1} as our main reasoning model. It is a
$14$B-parameter dense decoder-only transformer in the Qwen2.5 architecture~\citep{qwen2025qwen25} with $48$ layers, hidden dimension
$5{,}120$, $40$ attention heads with $8$ key-value heads, and a vocabulary of $152{,}064$ tokens.
We choose an open-weight model because most of our experiments require access to the model's internals rather than just its final outputs.

In each round, we present the LLMs with the payoff matrix and interaction history from the most recent ten rounds before asking them to choose their next action. The full prompt is provided in the Appendix (Box~\ref{box:game-prompt}). To account for the player's limited memory capacity, we use a rolling window of the ten most recent rounds rather than the entire game history. Each game lasts thirty rounds, but the LLM is unaware of this; the prompt simply indicates that there is a good chance the game will continue to the next round. We provide an example of a complete game in Figure~\ref{fig:gameplay}.

\begin{figure*}[t]
\centering
\includegraphics[width=\textwidth]{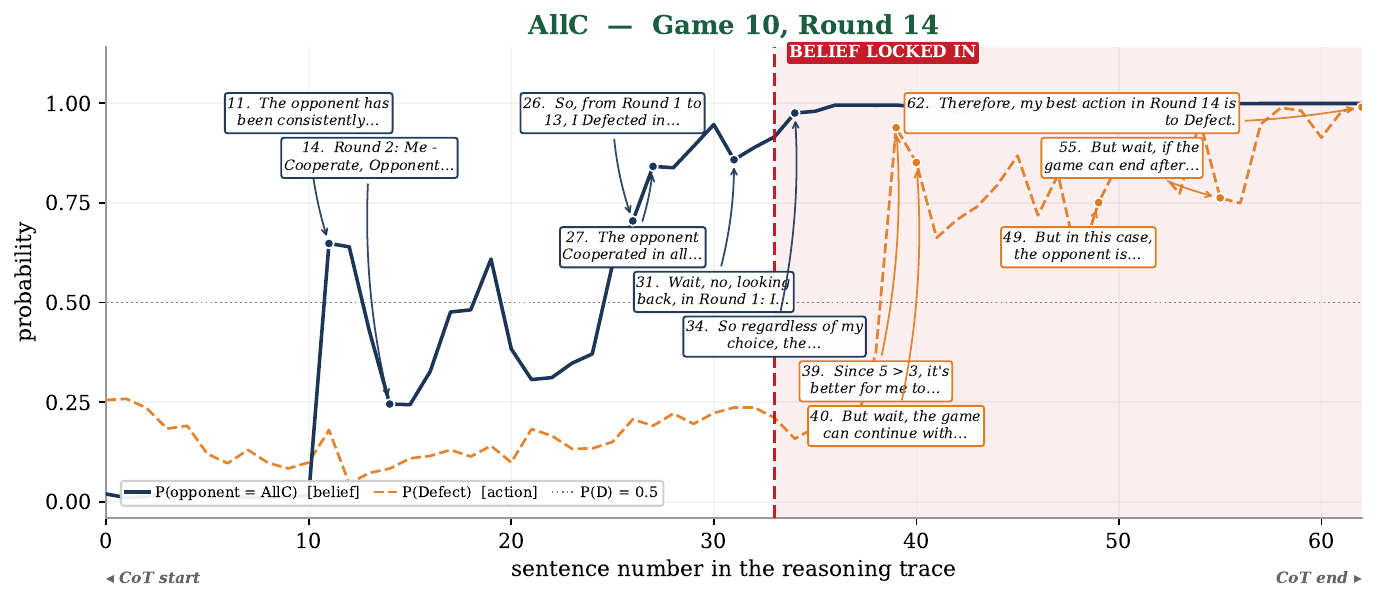}
\caption{A snapshot of a game against \textsc{AllC} at round 14, where we read $P(\text{opponent} =
\textsc{AllC})$ (navy curve) and $P(\mathrm{Defect})$ (dashed amber curve) after each sentence of the trace. Once belief is locked onto the true opponent class, the decision to defect surges afterward.}
\label{fig:belief-action}
\end{figure*}

\paragraph{Opponents.} We deploy five fixed strategy archetypes for the opponent. Three are stateless: \textsc{AllC} plays \Cop with
probability $1$, \textsc{AllD} plays \Def with probability $1$, and \textsc{Random} plays \Cop with
probability $0.5$. Two are reactive to the model's behavior: \textsc{TFT} opens with \Cop and mirrors the model's previous action, and \textsc{Grim} plays \Cop until the model defects once and then plays \Def forever.
Because the name of the game or the opponent's true strategy class is never revealed in the prompt, we hypothesize that the model should construct a mental model of the opponent from the provided context, such as payoff structure or observed history. It is in our interest to investigate how the LLM processes such evidence and applies strategic reasoning to reach its final decision.

\section{How does thinking affect decision-making?}
\label{sec:thinking}

CoTs are useful insofar as the generated traces are used in the actual reasoning that influences the corresponding decision~\citep{wei2022chain}. Earlier work~\citep{lanham2023measuring, chen2025reasoning} has shown that decisions in reasoning models are consolidated early in the trace, after which the remaining text has little causal bearing on the final response.
This discrepancy implies that CoT texts could be nothing more than an unfaithful, post-hoc rationalization masquerading as the model's genuine reasoning.

In this section, we revisit the above assumption by investigating how the LLM's internal judgment and decisions evolve sentence by sentence over their reasoning traces.
\subsection{Methodology}
\paragraph{Chunking and asking.}
We first split each round's CoT into $K$ chunks $\{c_1, \dots, c_K\}$ and, for each index $k \in \{0, 1, \dots, K\}$, prompt the LLM to verbalize its verdict using only the partial trace
\begin{equation}
    \mathbf{c}_{\leq k} = c_1 \oplus \cdots \oplus c_k,
\end{equation}
which is concatenated up to the first $k$ sentences of the full trace.
Here, we consider two question prompts $q_{\text{opp}}$ and $q_{\text{act}}$ (see Box~\ref{box:action-probe} and Box~\ref{box:belief-probe} in the Appendix) to probe the two main variables of interest: the perceived identity of the opponent (i.e., belief) and the decision to cooperate or defect (i.e., action). Formatting them as multiple-choice questions, we can find the next token probabilities for each answer option by unembeding the residual stream at the final layer and recording the logits. We then take the softmax over the single-letter tokens to get $P(\mathrm{\text{act} = \Def})$ and $P(\text{opp} =\textsc{AllC})$.
If the CoT is required to make the final decision, we can expect these probabilities to fluctuate as the reasoning chunks accumulate.

\paragraph{Sentence Ablation.} To better understand the (relative) significance of each sentence in the CoT, we also investigate how removing the chunk $c_i$ from the original trace affects the subsequent token logits. For each downstream chunk $c_j$, we calculate the KL divergence between two next-token distributions at each index $j > i$: the baseline trace $\mathbf{c}_{\leq K}$ and the masked trace
\begin{equation}
\mathbf{c}_{-i} = c_1 \oplus \cdots \oplus c_{i-1} \oplus c_{i+1} \oplus \cdots \oplus c_K.    
\end{equation}
Averaging over all $c_j$'s token positions yields how much removing sentence $i$ affects the model's predictions across sentence $j$. Using this measure, we can connect pairs of sentences with strong dependence and iterate across the full trajectory (see Appendix~\ref{app:causal} for further details).

\subsection{Case Study}
\label{sec:case-study}

To provide intuition, we start with a small-scale investigation of the LLM's thinking process in the middle of a game (round 14) against \textsc{AllC}.
Figure~\ref{fig:belief-action} tracks the two probabilities of interest using direct question prompts. Initially, the model is biased toward cooperation and spends the first ten sentences repeating the game prompt, during which it has not yet developed a mental model of the opponent.
As the LLM attempts to generalize the patterns from the opponent's history around $c_{11}$ (``\texttt{The opponent has been consistently choosing Cooperate in almost all rounds...}''), we see $P(\text{opp} =\textsc{AllC})$ rising above chance, indicating the model's ability to update their internal world model in response to new information. Moreover, removing this sentence has a causal effect on the model's predictions for subsequent sentences in which the opponent's belief eventually settles. The complete trace from the sentence ablation experiment is provided in the Appendix (Figure~\ref{fig:causal-chain})

Note that the belief is still fickle at the early stage. With multiple backtracking traces (''\texttt{Wait,...}''), the LLM self-corrects its earlier misguided assumption that the opponent cooperates in most but not all rounds by subsequently performing several independent readings of the history. Such indecisiveness is reflected in the erratic evolution of $P(\text{opp} =\textsc{AllC})$ before stabilizing at the midpoint of the trace around $c_{34}$ (``\texttt{So regardless of my choice, the opponent has always Cooperated.}``). After the LLM has confidently determined the opponent's identity, it takes as few as five chunks to translate that knowledge into a decision to defect, causing $P(\mathrm{\text{act} = \Def})$ to abruptly rise and remain in the defection regime until the end of the trace.

\subsection{Results} \label{sec:thinking-results}

The first pass results in Section~\ref{sec:case-study} lends credence to the notion that (a) the CoT can influence the LLM's behavior, and (b) the LLM can store internal representations of the game's latent structures and use them to guide the final action. To quantify these claims, we repeat the aforementioned analysis on 250 independent games (50 for each opponent).

\subsubsection{CoT allows for belief and decision updating.}

LLMs form hypotheses and iteratively revise them in light of new information. Although the prompt contains all of the necessary information for inference, their initial guess about the opponent or the strategy prior to CoT is almost always inadequate. Thinking out loud generally improves the opponent's identification accuracy and the probability of landing on the optimal action (see Table~\ref{tab:c1} in the Appendix).

\paragraph{Anatomy of reasoning.} Applying the sentence ablation protocol to the entire game population, we find that most reasoning traces follow the same qualitative shape.
Before making a decision, the model performs three core computations: \textbf{(i)} reviewing the history log, \textbf{(ii)} formulating and confirming the opponent's hypothesis, and \textbf{(iii)} calculating the expected payoff. The ambiguity with which the opponent can be identified---that is, the number of possible policies that yield the same behavior as presented in the history log---determines the order and quantity of these traces.

We present a few interesting patterns of the model's reasoning process:
\begin{itemize}
    \item For easy-to-verify games, such as those against \textsc{AllC} and \textsc{AllD}, the model allocates around $31\%$ of the total chains to identifying the opponent's strategy. In games with more ambiguity, this amount is cut in half, and more tokens (about $46\%$ and $50\%$ by mid and late game) are reallocated to backtracking and reciting the round-by-round history. This transition shows the LLMs expressing more uncertainty about their opponent.
    \item Because reasoning in ambiguous games is crowded out by history recap sentences, other computations are deferred late within the trace, which is exacerbated as more rounds unfold. The end result is rash decision-making, where the LLM retracts an already correct belief. As an example, out of all the \textsc{TFT} rounds that conclude with the incorrect opponent at the end of the trace, $94.6\%$ had the correct answer earlier in the trace but backtracked. A case study of an individual round is provided in the Appendix (Figure~\ref{fig:tft}).
\item Sentences that express confusion gradually deteriorate as the rounds progress. Over time, the model's confidence increases regardless of the validity of the belief. This premature commitment implies that if it converges on the incorrect representation about the world, it is unlikely to reassess, inevitably becoming stuck on the suboptimal move.
\end{itemize}

\begin{figure}
    \centering
    \includegraphics[width=\columnwidth]{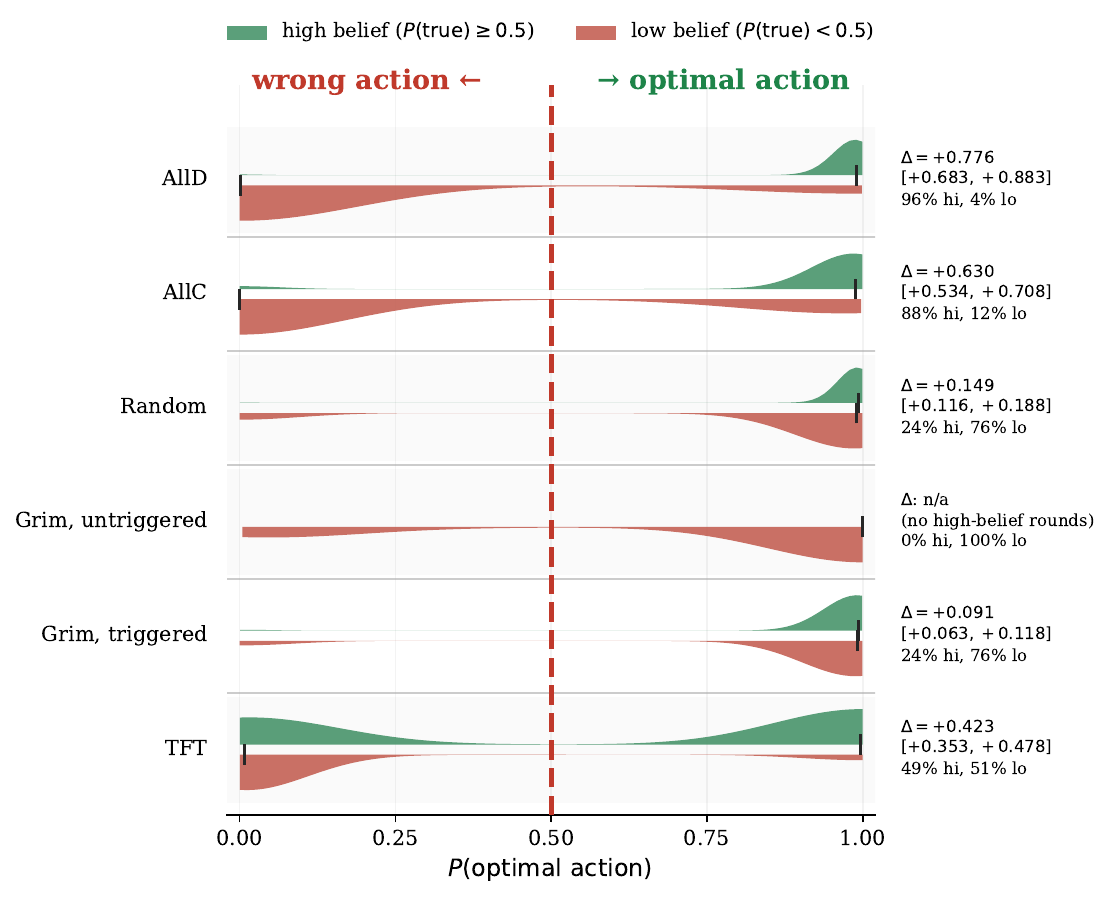}
    \caption{Belief-action coupling by opponent condition, split by final belief. The rows divide the distribution of $P(\text{optimal action})$ for each round into high-belief ($P(\text{opp = true}) \geq 0.5$, green) and low-belief ($<0.5$, rose) halves. The vertical tick on each half-violin marks its median. $\Delta$ is the gap in mean $P(\text{optimal action})$ between the high- and low-belief halves, with $95\%$ CIs from game-level bootstrap; percentages give each half's share of the condition's rounds.}
\label{fig:belief-action-splitviolin}
\end{figure}

\paragraph{Belief leads action within the trace.} The formation of the opponent's belief generally occurs before the commitment to cooperate or defect. To demonstrate this, we treat the belief and action probe trajectories within each trace as a bivariate time series and fit a cross-lagged (Granger) model on their first differences. We compare the standardized belief-to-action coefficient $\beta_{ba}$ against the reverse $\beta_{ab}$.

Our finding shows that $\beta_{ba} > \beta_{ab}$ at every lag and for every opponent. The standardized gap is smaller for opponents it finds difficult to determine ($0.084$ for \textsc{TFT}, $0.032$ for \textsc{Grim}, and $0.049$ for \textsc{Random}), and it is largest for opponents it consistently pins down ($0.168$ for \textsc{AllC} and $0.126$ for \textsc{AllD}). Intuitively, when the model can determine who it is dealing with, it commits to a belief about the opponent early in the trace, and the action follows suit; otherwise, the belief remains noisy, providing little signal for the action to track. We also find that the boundary where the model's readout belief probabilities do not deviate from their final value occurs at a median $29\%, 40\%, 67\%, 82\%$, and $88\%$ of the trace for \textsc{AllD}, \textsc{AllC}, \textsc{Random}, \textsc{Grim}, and \textsc{TFT}, while actions are always committed near the end (at a median $98\%$) of the trace.

\subsubsection{Belief shapes action.}\label{subsec:belief-action}

The LLM's final action is correlated with its belief's content and the earliest point with which it stabilizes in the trace (i.e., how soon the model becomes sufficiently confident).
Figure~\ref{fig:belief-action-splitviolin} depicts this relationship by dividing each opponent's round distributions by the strength of their beliefs and comparing the resulting $P(\text{optimal action})$ splits.

We observe evidence of a belief-action coupling across most games, which suggests that accurate identification of the opponent helps the model reach the game-theoretically optimal action. However, identification is not strictly necessary. It is most important when the best response is to cooperate, and less important when the best response is to defect, which the model already prefers in the presence of uncertainty. Against \textsc{Random}, for instance, the model identifies the opponent poorly yet still plays near-optimally (ID accuracy $0.26$ versus $P(\text{optimal})=0.87$ at $k{=}K$; Table~\ref{tab:c1}), since constant defection is the uncertainty-robust response. As expected, the model has high confidence in the belief of \textsc{AllD} and \textsc{AllC}, so it plays \Def with high certainty for most rounds. The untriggered \textsc{Grim}, by definition, occurs early in the game, when the LLM has not tried an alternate strategy to gather information, and thus its prior belief (which is to play \Cop) does not correlate with the best action. Almost half of the \textsc{TFT} rounds end up with the incorrect model about the opponent, but those with the correct belief learn to reciprocate and play \Cop accordingly.

\begin{figure*}
    \centering
    \includegraphics[width=\textwidth]{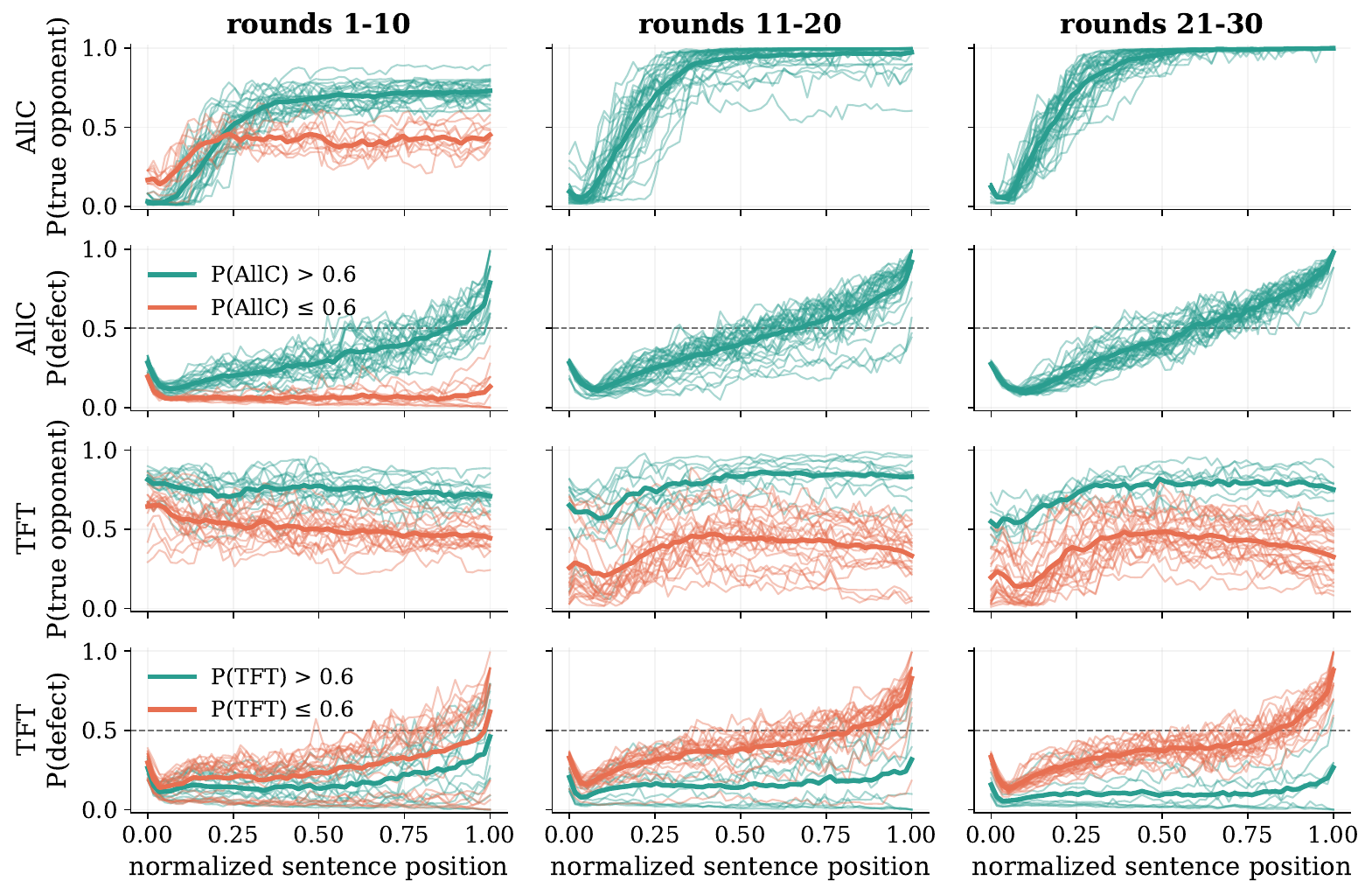}
\caption{Belief ($P(\text{opp = true})$ and action ($P(\text{act}=\Def)$ trajectories within the reasoning trace for \textsc{AllC} and \textsc{TFT}, faceted by round bin. Games are split within each bin by whether their opponent probe ends the trace above (teal) or at/below (coral) 0.6. Thin lines are individual games; thick lines are per-cluster means.}
\label{fig:evolution}
\end{figure*}

We further illustrate this nuanced dynamic in Figure~\ref{fig:evolution}, which inspects the probed probabilities $P(\text{opp = true})$ and $P(\text{act} = \Def)$ in the reasoning trace during (early, mid, late) games against \textsc{AllC} and \textsc{TFT}.

\paragraph{Accurate beliefs improve performance.}
In the early rounds against \textsc{AllC}, the LLM sometimes misidentifies the opponent and reasons its way to cooperation ($P(\text{act=}\Def) \approx 0.1$--$0.2$), which is not the optimal strategy. We expect this noisy performance, as reasoning models do computations stochastically and may commit to the wrong reasoning path, particularly when the decisions involved are complex. Ultimately, the model learns to self-correct. The true opponent's identity recovered by rounds 11--20, resulting in a gradual increase in $P(\text{act=}\Def)$ throughout the trace. This consolidated knowledge of the opponent, as well as the consequent commitment to defect, persists for the remainder of the game.

\paragraph{Wrong beliefs diminish performance.}
Because CoT-driven revision does not always lead to the truth, we can expect worse performance in rounds with misclassified opponents. We should note, however, that a low probability of the true type is not always an inference error. Since each round only shows a truncated ten-round history, several opponent types can be observationally equivalent (see Section~\ref{sec:thinking-results}).
As reflected in Figure~\ref{fig:evolution}, we see the games against \textsc{TFT} bifurcate into two categories with time: those where the LLM can identify the opponent (teal) and those where it cannot (coral). As mutual defects accumulate in the log, the model increasingly misinterprets the opponent as \textsc{AllD}. This causes $P(\text{act = \Def})$ to climb past 0.5, leading to suboptimal retaliation. Remarkably, we find the same bias towards defection against \textsc{Random} and \textsc{Grim}, since the LLM also struggles to identify the opponent in these situations (see Figure~\ref{fig:belief-action-splitviolin} and Appendix~\ref{app:perf}). However, the bias is only detrimental against \textsc{Grim}, where defection ultimately results in irreversible retaliation. In contrast to \textsc{Random}, defection happens to coincide with optimal play, so poor identification of the opponent does not matter.
Looking ahead, our findings point to the need for human supervision and/or guardrails to mitigate these systemic vulnerabilities.

\begin{figure*}
    \centering
\includegraphics[width=\textwidth]{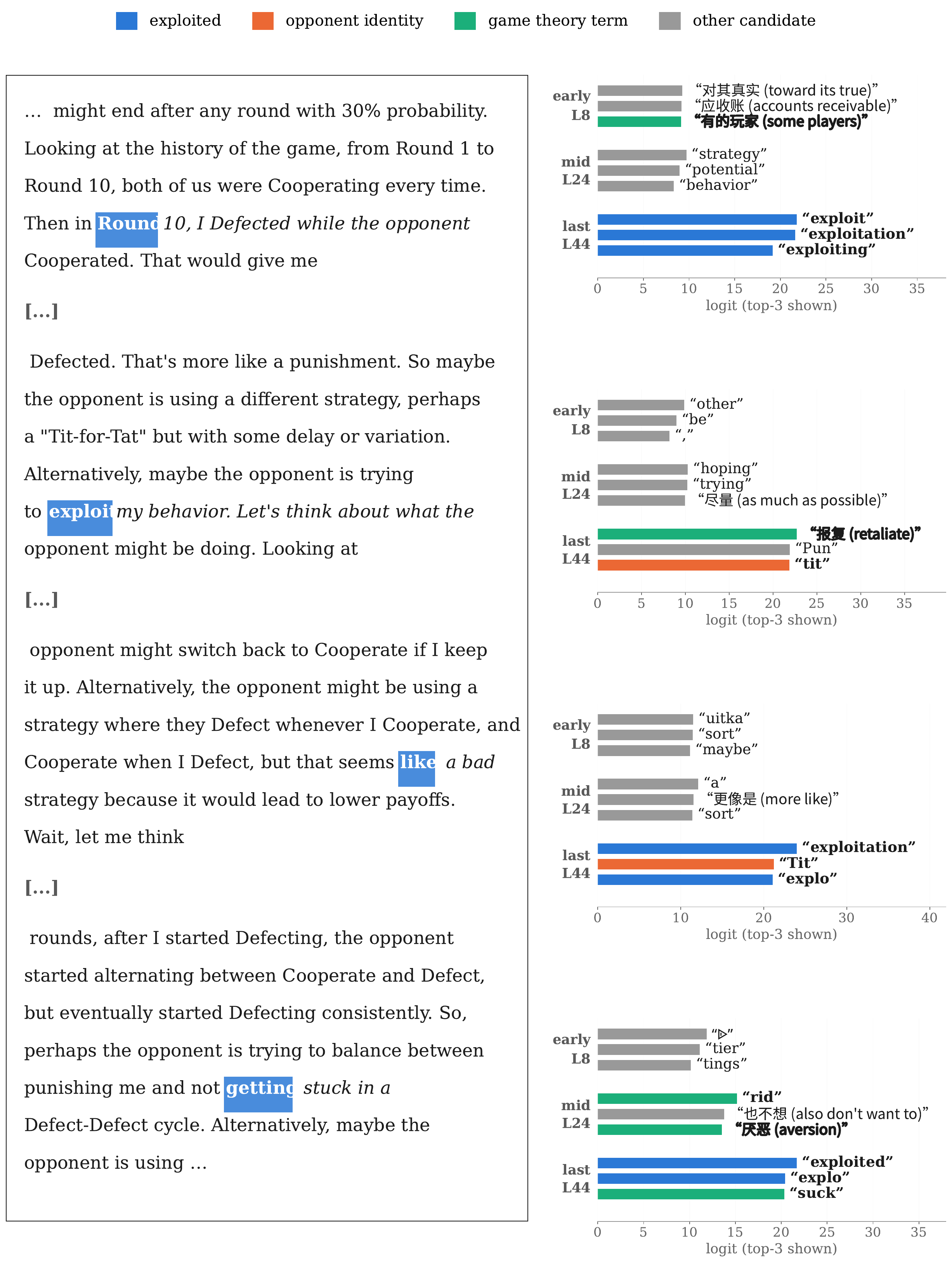}
    \caption{Four readouts of $\mathrm{lens}(h_\ell)$ across TFT round 17's reasoning trace. Left: CoT excerpts, with the readout position boxed. Right: top-3 candidates at an early ($\ell{=}8$), middle ($\ell{=}24$), and the final captured ($\ell{=}44$) layer for that position. Row 1: 605 tokens before ``\texttt{exploit}'' is ever articulated. Row 2: 1 token after the literal mention of ``\texttt{exploit}''. Rows 3--4: after the round's only appearance of ``\texttt{exploit}'', at $29\%$ and $56\%$ through the trace respectively.}
    \label{fig:exploit-still-thinking}
\end{figure*}

\section{Reading the intent beneath the CoT}

Our analysis in Section~\ref{sec:thinking} assumes that any inferences drawn by the LLM about the game are externalized into the CoT. However, it is possible that the model relies on intermediate ideas that do not manifest in the reasoning process or are phrased using a different vocabulary than the question.
According to recent research on global workspace theory in LLMs~\citep{gurnee2026verbalizablerepresentationsformglobal,chalmers2026jacobian}, these hidden variables can emerge in the model's ``workspace'' layers during the forward pass and can be decoded from a linear readout of intermediate activations.

\subsection{Methodology}

\paragraph{Jacobian lens.} The Jacobian lens~\citep[J-lens;][]{gurnee2026verbalizablerepresentationsformglobal}, approximates the first-order causal effect of activations on the model's output tokens. It uses a Jacobian to account for changes in coordinate systems from the intermediate layers to the final layer to improve comprehension of the readouts in intermediate layers. 
For every layer $\ell$, we fit the linear map
\begin{equation}
    J_\ell = \mathbb{E}_{t,\,t' \ge t}\left[\frac{\partial h_{\mathrm{final},t'}}{\partial h_{\ell,t}}\right],
\end{equation}
by backpropagating from the final-layer residual stream down to a layer-$\ell$ activation and averaging over token positions and a corpus of prompts.
Applying the unembedding matrix to the corrected activation provides a token-level ranked readout of what the model is poised to verbalize from any layers at every token position in the CoT. This reveals the intermediate phases of the model's internal reasoning, which may include ideas that the model internally conceives but never articulates.

\subsection{Case Study}
We first examine round 17 of a game against \textsc{TFT}, where the LLM prematurely commits to the incorrect belief, as alluded to in Section~\ref{subsec:belief-action}.
Perusing this round's CoT, we observe that the model initially considered the reciprocity hypothesis (\texttt{using a Tit-for-Tat strategy?}), but during the first $6\%$ of the trace, it became preoccupied with the prospect of being punished (\texttt{trying to punish me}) and exploited (\texttt{trying to exploit my behavior}). This diversion compels the LLM to confabulate and refute its initial belief of the opponent, leading to the irrational action of playing \Def in this round.

\begin{figure}[b]
    \centering
    \includegraphics[width=\columnwidth]{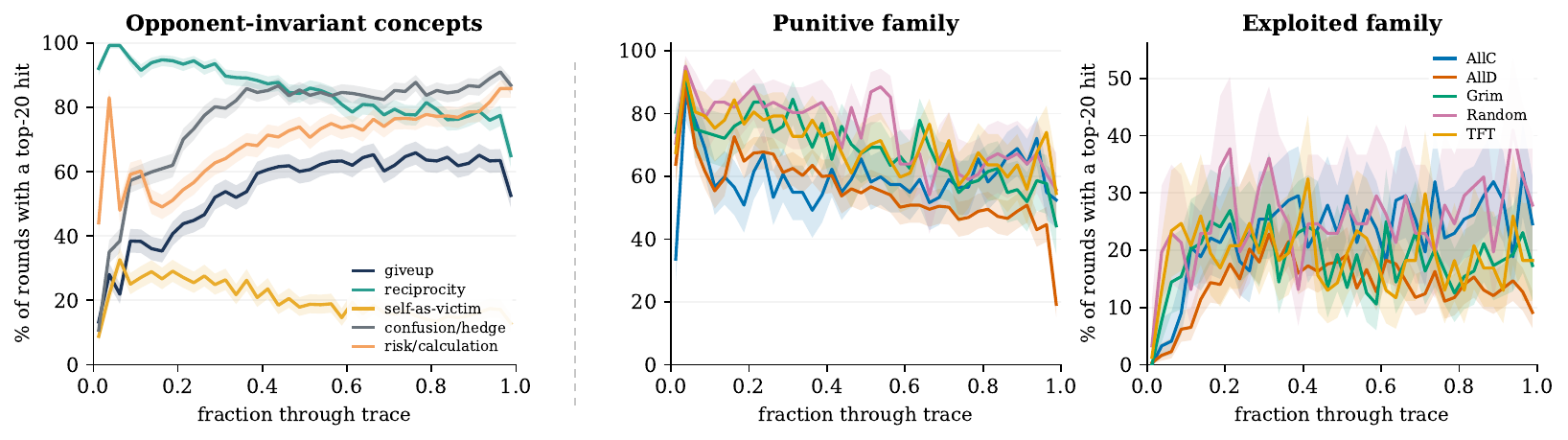}
    \caption{Presence rate of concept families across the reasoning trace (band: Wilson $95\%$ CI). \textbf{Opponent-invariant concepts:} five vocabulary families whose trajectory shape agrees across opponents (mean pairwise Pearson $r=0.65$--$0.91$), pooled into one curve per family. \textbf{Punitive family, Exploited family:} the two families whose shape depends on the opponent ($r=0.55, 0.40$).}
    \label{fig:silent}
\end{figure}

We apply the J-lens to each token position in the CoT and record the top lens tokens. Figure~\ref{fig:exploit-still-thinking} annotates the CoT transcript with the J-lens readouts across layers.
At about $3\%$ through the trace, the lens begin to flag \texttt{exploit}, \texttt{exploitation}, and \texttt{exploiting} as the model tries to recollect the history and generalize the opponent's pattern. It should be noted that these labels emerge $605$ tokens before the exploit frame is directly mentioned in the CoT. LLM's nonverbal reasoning precedes its verbal dispositions.

\begin{CJK}{UTF8}{gbsn}
Another notable finding is that the model can reason over semantic neighborhoods of the same concept, both cross-lingually (e.g., \texttt{报复}, which means to retaliate in Chinese) and morphologically (e.g., \texttt{Pun}, a truncated variant of punish). As the model is weighing whether it is ``\texttt{getting stuck in a Defect-Defect cycle}'', the lens surface \texttt{exploited}, \texttt{explo}, \texttt{suck}, 厌恶 (aversion) at the mid and final layers. Moreover, the LLM appears to recognize that its environment pertains to economic games, as evidenced by the recurring appearances of \texttt{strategy}, \texttt{有的玩家} (some players), and \texttt{behavior} in the early and mid layers across the trace. 
\end{CJK}

\subsection{Results}
We now extend the above corpus inspection protocol to 250 independent games (50 for each opponent).
For each round, we divide the generated trace into $40$ equal-fraction bins and record J-lens' top-$20$ candidates for each token position. Figure~\ref{fig:silent} categorizes the most frequent readouts into seven vocabulary groups (see Table~\ref{tab:word-families} in the Appendix).

\paragraph{Five of the seven concept families are surfaced regardless of the opponent.} Most concepts in the model's internal layers take on sufficiently similar shapes across games (i.e., they are opponent invariant; see the first panel of Figure~\ref{fig:silent}). These surfaced tokens can provide us with an algorithmic intuition for how the LLM makes decisions in a repeated normal-form game.
With the exception of \textsc{reciprocity}, which syntactically aligns with one of the opponent's archetype names, most of these readouts do not appear in the actual CoT of the round.

We find that the LLM first considered the possibility of being taken advantage of via \textsc{self-as-victim} in the beginning of the CoT but set aside this frame as reasoning progressed.
In Section \ref{sec:thinking-results}, we established that the CoT in ambiguous games are crowded with hedging and history-recap language. Such reasoning patterns are consistent with
the emergence of \textsc{risk}, \textsc{hedge}, and eventually \textsc{giveup} token families in the J-lens toward the end of the trace.

\paragraph{Two concept families track each opponent's game-theoretic structure.} \textsc{punitive} and \textsc{exploited} have the weakest cross-opponent shape agreement ($r=0.55$ and $r=0.40$, versus $0.65$--$0.91$ for the others). We perform a Kruskal-Wallis test (see Figure~\ref{fig:silent-kw} in the Appendix), which reveals that both families reject the null hypothesis of no opponent effect (\textsc{punitive}: $H(4){=}108.7$, $p<0.001$; \textsc{exploited}: $H(4){=}68.5$, $p<0.001$). The second and third panels of Figure~\ref{fig:silent} display the per-opponent detected rate over the CoT, pooled across all rounds.

\textsc{exploited} is the least common (detected in a mean of $13.9\%$ bins per round) in games against \textsc{AllD}, where defecting is the optimal strategy and mutual defection leaves no one to exploit. However, it is highest against its naive, exploitable counterparts: \textsc{Random} ($25.4\%$) and \textsc{AllC} ($22.7\%$).
Since the appearance of \textsc{punitive} depends on whether the opponent's strategy is reactive to the model's actions, it is ranked lowest when compared to the two static archetypes \textsc{AllC} ($59.4\%$) and \textsc{AllD} ($55.8\%$). As expected, it ranks relatively high when compared to \textsc{Random} ($74.2\%$), \textsc{TFT} ($70.4\%$), and \textsc{Grim} ($68.4\%$), whose behavior is unpredictable enough to provoke thoughts of being punished.

\section{Discussion}
In this paper, we investigated how CoT reasoning traces affect the performance of LLM agents playing an iterative Prisoner's Dilemma.
Based on three levels of mechanistic evidence (logit probes, sentence ablation, and activation-level concept readouts), our findings provide a consistent picture of the LLM's decision-making process in a repeated normal-form game.
\textbf{First,} the model can retain and revise valuable latent game information (i.e., belief) over time, which aids in the selection of the final decisions (i.e., action). \textbf{Second,} the belief-action coupling is shown to be persistent: once its mental model of the game has stabilized, the LLM rarely revisits it, which can lead to suboptimal decisions that compound with time.
\textbf{Third,} CoT can be viewed as a partial account of the reasoning process. Upon examining the activations, we find that the model sometimes uses internal, nonverbalized concepts to inform its reasoning throughout the CoT.
Because our explanations are based on convergent lines of evidence, we think of internal representations as guiding (leading and shaping) the decision rather than being the sole causal factor.

\section*{Limitations}
Looking forward, our work suggests concrete research directions. First, we only look at an open-weight LLM of moderate size, which raises the question of whether significantly larger models behave similarly. A prior survey~\citep{wang2025safetylargereasoningmodels} indicates that scale alone cannot resolve safety failures, but future research should investigate this possibility. Second, our study only focuses on fixed, non-adaptive opponents. To accommodate the growing number of multiagent systems, our pipeline can be extended to incorporate LLM-vs-LLM play.
Another intriguing follow-up is to carry out an intervention experiment in an attempt to reverse the lock-in behavior, such as ablating the concept families from the residual stream at the point in the trace where belief solidifies.

\section*{Author Contributions}
Chaipat Tirapongprasert led the project, wrote the paper, and carried out all experimental work. Rajarshi Das and Parijat Dube provided feedback and guidance throughout the project.

\section*{Acknowledgments}
This work was supported in part by Columbia University's Department of Statistics.
C.T. thanks Zehao Peng for insightful discussions, which helped shape some of our approach. All the experiments were run on NCSA DeltaAI at UIUC under the Advanced Cyberinfrastructure Coordination Ecosystem: Services \& Support (ACCESS) program.

\bibliography{custom}

\appendix
\section{Payoff Matrix and Prompts}\label{app:ipd}

\paragraph{Payoff matrix.} We instantiate the payoff matrix for iterated prisoner's dilemma as $T = 5$ (temptation), $R = 3$ (reward), $P = 1$ (punishment), $S = 0$ (sucker). The ordering $T > R > P > S$ makes defection the dominant action in the one-shot game while mutual cooperation maximizes joint welfare \citep{rapoport1965prisoner,axelrod1981evolution}. Table~\ref{tab:payoff-matrix} gives the payoff to each player for
every combination of actions, with the model as the row player and cells written as (model's payoff, opponent's payoff).

\begin{table}[h]
\centering
\small
\begin{tabular}{llcc}
\toprule
& & \multicolumn{2}{c}{Opponent} \\
\cmidrule(lr){3-4}
& & \Cop & \Def \\
\midrule
\multirow{2}{*}{Model} & \Cop & $R,R$ \;(3, 3) & $S,T$ \;(0, 5) \\
& \Def & $T,S$ \;(5, 0) & $P,P$ \;(1, 1) \\
\bottomrule
\end{tabular}
\caption{Payoff matrix for the iterated prisoner's dilemma game used throughout the paper.}
\label{tab:payoff-matrix}
\end{table}

\paragraph{Game prompt.} Every round, the model is shown the payoff matrix, the continuation
probability, and up to the ten most recent rounds of history, then asked to commit to an action
in a fixed format.

\refstepcounter{promptboxctr}
\begin{promptbox}{IPD game prompt}
\label{box:game-prompt}
You are playing a repeated two-player game against a fixed opponent. Each round, you and your opponent each simultaneously choose one of two actions: Cooperate or Defect.\\[4pt]
Payoffs this round, depending on what you and your opponent choose:\\
- You Cooperate, Opponent Cooperates: you get 3, opponent gets 3\\
- You Cooperate, Opponent Defects: you get 0, opponent gets 5\\
- You Defect, Opponent Cooperates: you get 5, opponent gets 0\\
- You Defect, Opponent Defects: you get 1, opponent gets 1\\[4pt]
After this round, there is a 70\% chance the game continues for at least one more round, and otherwise the game ends.\\[4pt]
Here is what happened in the most recent rounds:\\
Round \textnormal{\textit{i}}: You chose \textnormal{\textit{Cooperate$\mid$Defect}}, Opponent chose \textnormal{\textit{Cooperate$\mid$Defect}}.\\
\textnormal{\textit{$\vdots$ (up to 10 most recent rounds; ``This is the first round of the game.'' if none)}}\\[4pt]
It is now Round \textnormal{\textit{n}}. Think step by step about what your opponent is likely to do and what action best serves your long-run interest. When you have decided, end your response with a line in exactly this format (and nothing after it):\\FINAL ACTION: Cooperate\\
or\\
FINAL ACTION: Defect
\end{promptbox}

\paragraph{Probe prompts.} At each chunk boundary $k$ (Section~\ref{sec:thinking}), we truncate
the trace after $\mathbf{c}_{\leq k}$, close the \thinktag block, and force one of the two continuations below. Both are appended verbatim after the truncated trace, so the model's response is contingent on the reasoning it has generated thus far. We read next-token logits at the final position: a binary Cooperate/Defect
softmax for the action probe (Box~\ref{box:action-probe}), and a five-way softmax over the option letters for the belief probe (Box~\ref{box:belief-probe}).

\begin{figure*}
    \centering
    \includegraphics[width=\textwidth]{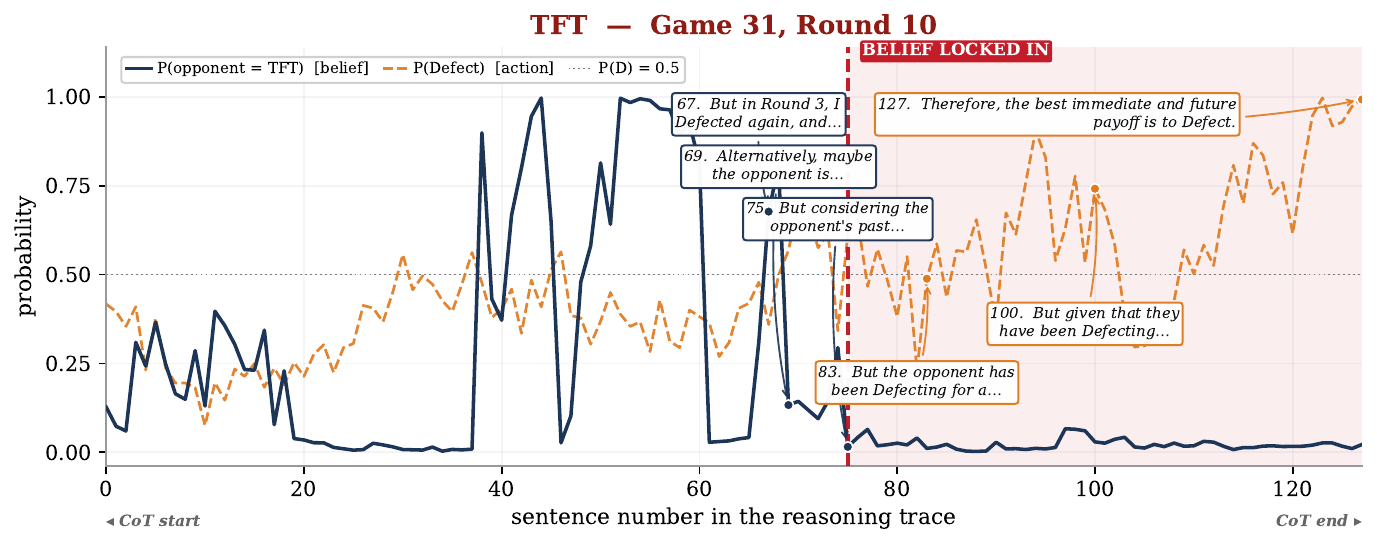}
    \caption{Sentence-level traces of $P(\text{opponent class})$ (navy) and $P(\mathrm{Defect})$ (dashed amber) against \textsc{TFT}. An early belief spike is retracted mid-trace as later sentences revise the opponent estimate, demonstrating the model's shift away from the ground truth.}
    \label{fig:tft}
\end{figure*}

\begin{table*}[t]
\centering
\begin{tabular}{lcccc}
\toprule
Opponent & ID acc., $k{=}0$ & ID acc., $k{=}K$ & P(optimal), $k{=}0$ & P(optimal), $k{=}K$ \\
\midrule
\textsc{AllC}   & 0.017 & \textbf{0.878} & 0.271 & \textbf{0.833} \\
\textsc{AllD}   & 0.906 & 0.960          & 0.342 & \textbf{0.936} \\
\textsc{Random} & 0.000 & 0.261          & 0.320 & \textbf{0.868} \\
\textsc{Grim}   & 0.097 & 0.207          & 0.407 & \textbf{0.879} \\
\textsc{TFT}    & 0.493 & 0.514          & 0.713 & 0.352 \\
\bottomrule
\end{tabular}
\caption{Read out opponent-identification accuracy and probability of optimal game-theoretical action before reasoning ($k{=}0$) and after the complete trace ($k{=}K$). Bold $k{=}K$ values indicate progress toward ground truth.}
\label{tab:c1}
\end{table*}

\refstepcounter{promptboxctr}
\begin{promptbox}{Action probe}
\label{box:action-probe}
\textnormal{\textit{[$\cdots$ truncated trace $\mathbf{c}_{\leq k}$ $\cdots$]}}\\
\endthinktag\\[6pt]
FINAL ACTION:\ \textnormal{\textit{[read $P(\text{Cooperate})$ vs.\ $P(\text{Defect})$ here]}}
\end{promptbox}

\refstepcounter{promptboxctr}
\begin{promptbox}{Opponent-identity (belief) probe}
\label{box:belief-probe}
\textnormal{\textit{[$\cdots$ truncated trace $\mathbf{c}_{\leq k}$ $\cdots$]}}\\
\endthinktag\\[6pt]
The opponent's strategy is most likely:\\
(A) Always Cooperate\\
(B) Always Defect\\
(C) Tit-for-Tat\\
(D) Grim Trigger\\
(E) Random\\
Answer: (\textnormal{\textit{[read the 5-way softmax over A--E here]}}
\end{promptbox}

\section{Causal dependency between sentences}\label{app:causal}

In this section, we give more details about our causal ablation implementation. Recall that for a round with chunks $c_1,\dots,c_K$, we run one baseline forward pass over the entire trace and then one masked forward pass per chunk $i$ with $c_i$ ablated.
To calculate the sentence pairs' KL divergence $\mathrm{KL}(i\to j)$, we average the per-token KL divergence between the baseline's and the masked pass's next-token distributions over $c_j$'s tokens for each downstream chunk $j>i$. Because nearby sentences mechanically give larger $\mathrm{KL}$ than distant ones, we normalize each pair against $\overline{\mathrm{KL}}(d)$, the average $\mathrm{KL}(i\to j)$ across every pair in that same round at the same distance $d=j-i$.

\begin{figure*}[t]
    \centering
    \includegraphics[width=1.1\textwidth]{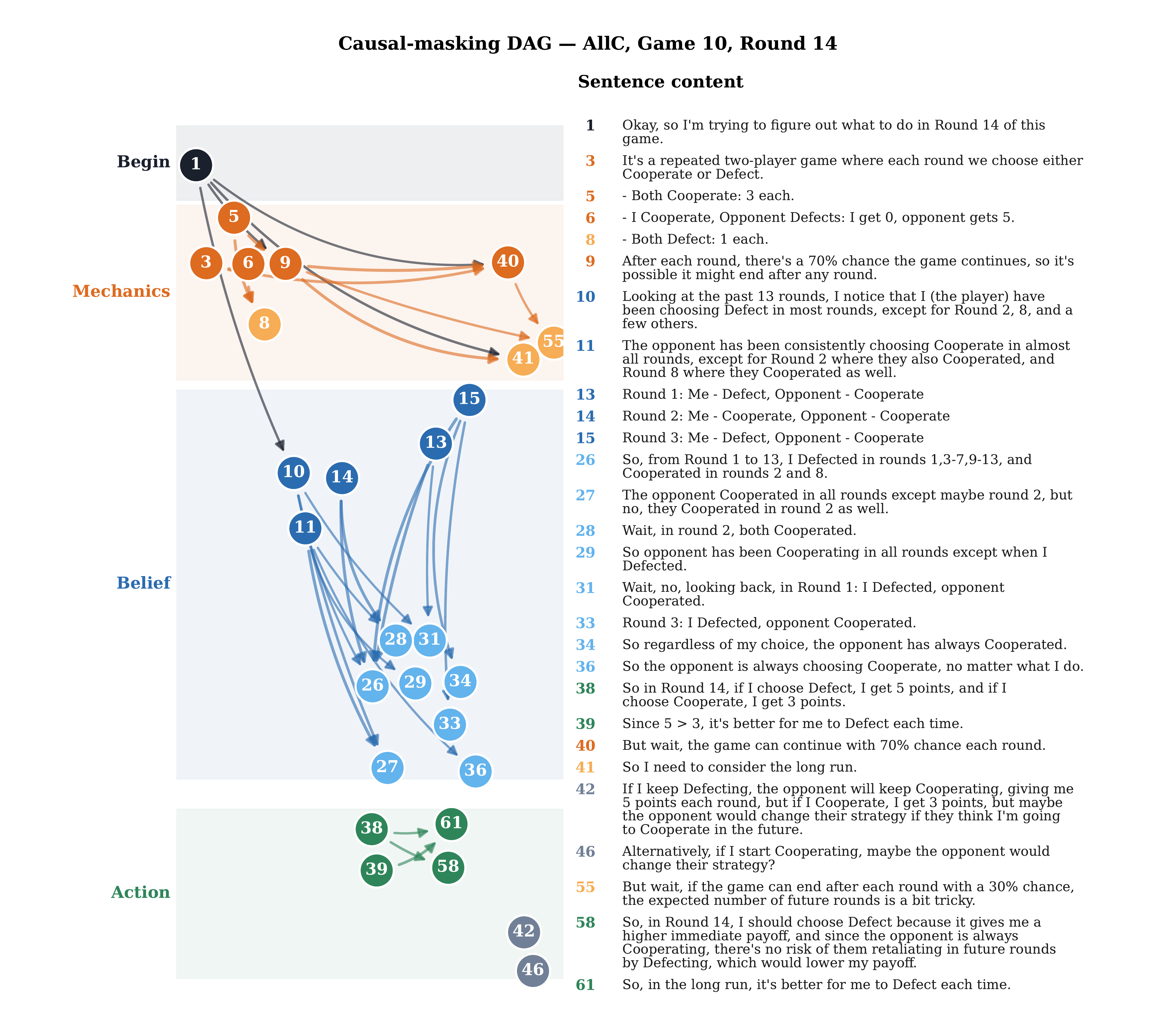}
    \caption{Causal reasoning chain for Round 14 against \textsc{AllC}. Belief-related chains follow a fan-out structure, whereas the action chain develops independently. The position of each node on the page is strictly cosmetic.}
    \label{fig:causal-chain}
\end{figure*}

\paragraph{Casual chain.} We define two sentences $c_i$ and $c_j$ as causally dependent if the sentence pair $\mathrm{KL}(i\to j)$ is at least $8\times$ larger than the distance-matched baseline $\overline{\mathrm{KL}}(d)$. Figure~\ref{fig:causal-chain} depicts the entire causal trajectory for Round 14 against \textsc{AllC}. We can see that the belief cluster is more dense and well-connected because the opponent's belief is updated by reading independently sufficient sections of the history log. Once the opponent's belief is confirmed, committing to an action can be carried out in a sequential reasoning chain.

\paragraph{Case study for retraction.} Figure~\ref{fig:tft} traces the failure mode at round 10 of a game against \textsc{TFT}. In the early traces, $P(\text{opp} = \textsc{TFT})$ oscillates violently, twice spiking to near $1.0$ before collapsing back toward $0$. During this period, $P(\mathrm{Defect})$ remains close to the $0.5$ decision boundary, indicating that the model has not yet committed to any action.
The belief returns to $1.0$ around sentence 50 and remains there until sentence 67 (''\texttt{But in Round 3, I Defected again, and\ldots}''), when the model revisits an old defection in the history log. Over the next few sentences, it talks itself out of the correct belief, and by sentence 75, $P(\text{opp} = \textsc{TFT})$ vanishes while $P(\mathrm{Defect})$ begins to rise.
We present this snapshot as an example of the premature-commitment dynamic, in which the model does not revisit an incorrect belief for the remainder of the round after it has stabilized.

\section{Addtional Population Analysis}\label{app:perf}

\paragraph{Evolution of beliefs and action.}
Table~\ref{tab:c1} displays the opponent-identification accuracy and probability of taking the game-theoretically optimal action before any reasoning has occurred ($k{=}0$, i.e., from the payoff matrix and history alone) versus after the complete reasoning trace ($k{=}K$). For \textsc{AllC}, \textsc{Random}, and \textsc{Grim}, reasoning is unambiguously useful. Both quantities increase significantly over the trace, confirming that the model begins with no opinion about the opponent's identity and gradually develops one during the reasoning process. \textsc{TFT} is the only condition where a naive, non-reasoning readout performs better. We discuss this failure mode in Section~\ref{sec:thinking-results}. Overall, the findings suggest that reasoning traces aid in the translation of beliefs (whether correct or incorrect) into actions.

\begin{figure}
    \centering
    \includegraphics[width=\columnwidth]{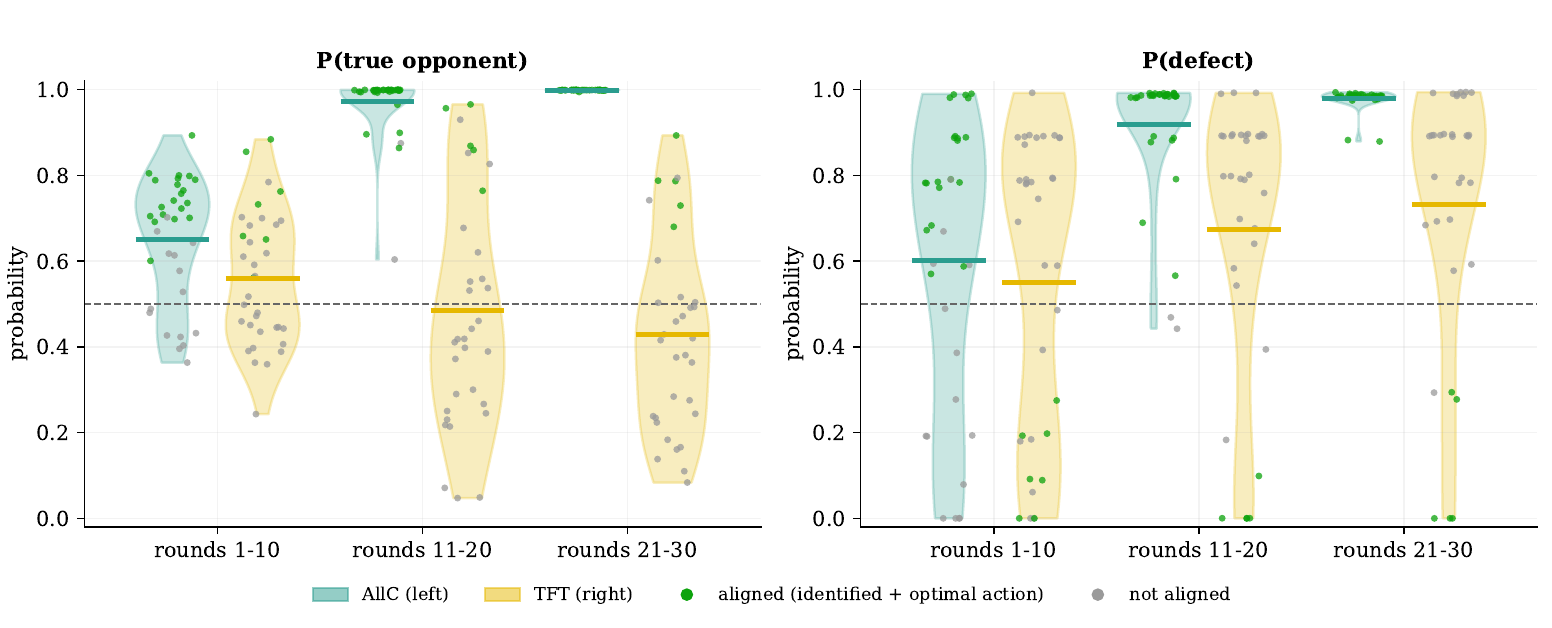}
    \caption{Side-by-side comparison of AllC and TFT (rounds 1-10, 11-20, and 21-30). Each violin is a per-game distribution (green = AllC, yellow = TFT); horizontal bars mark violin means. Left: $P(\text{opp = true})$ at round-end; right: $P(\text{act = \Def})$ at round-end. ``Aligned'' denotes games in which the LLMs identified the opponent more accurately than chance and consequently took the optimal strategy.}
    \label{fig:allc-tft}
\end{figure}

\paragraph{Game alignment}. Figure~\ref{fig:allc-tft} gives the per-game distributions for each round bin, colored by whether the game is aligned (i.e., the LLM identified the opponent and chose the correct action). This design aims to quantify how frequently an accurate belief leads to the game-theoretically optimal action.

Looking at the game trajectory against \textsc{AllC}, both defection and belief converge tightly toward $1$: mean $P(\text{opp = true})$ increases from $0.65$ in rounds 1–10 to $0.98$ by rounds 11–20 and $1.0$ by rounds 21–30. As expected, $P(\text{act} = \Def)$ closely tracks ($0.60 \to 0.92 \to 0.98$), with an increasing majority of games falling into the aligned cluster. In contrast, games against \textsc{TFT} exhibits the opposite characteristic, with its belief distribution becoming bimodal. This is seen in its mean drifting downward from $0.56$ in rounds 1–10 to $0.48$ in rounds 11–20 and $0.43$ in rounds 21–30. Because \textsc{TFT}'s optimal response is to cooperate, we see $P(\text{act} = \Def)$ climbing up from $0.55$ to $0.73$. This failure mode is described in Section~\ref{sec:thinking-results}.

\begin{figure}[htbp]
    \centering
    \begin{subfigure}[b]{\linewidth}
        \centering
        \includegraphics[width=\linewidth]{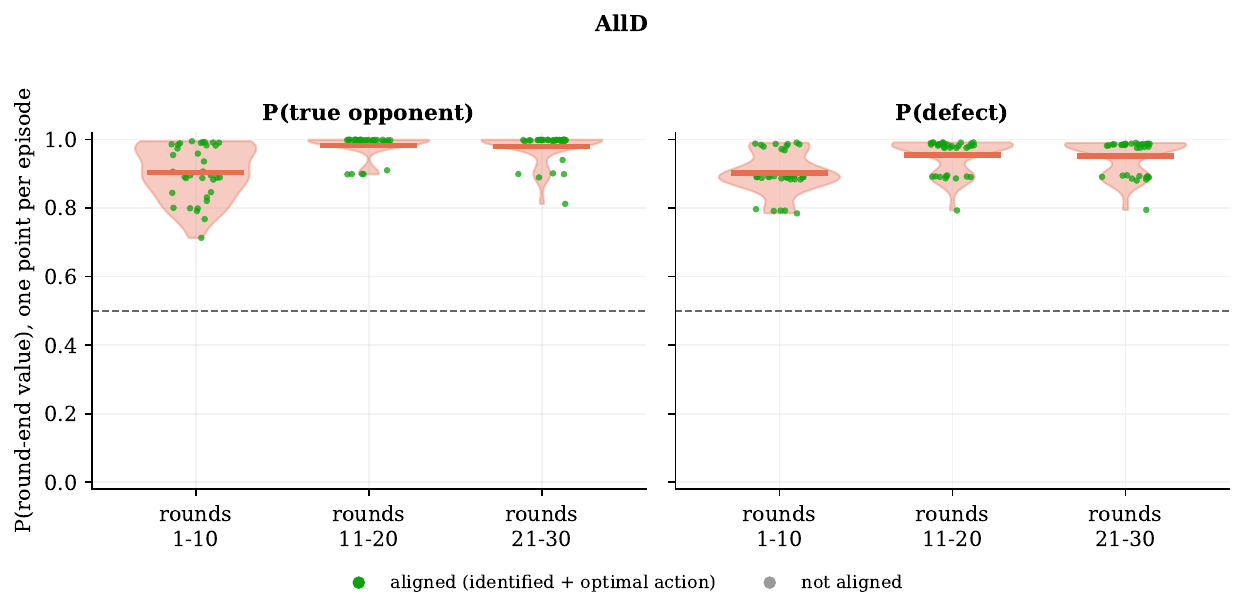}
        \label{fig:round_bin_violin_alld}
    \end{subfigure}
    \vspace{0.5em}
    \begin{subfigure}[b]{\linewidth}
        \centering
        \includegraphics[width=\linewidth]{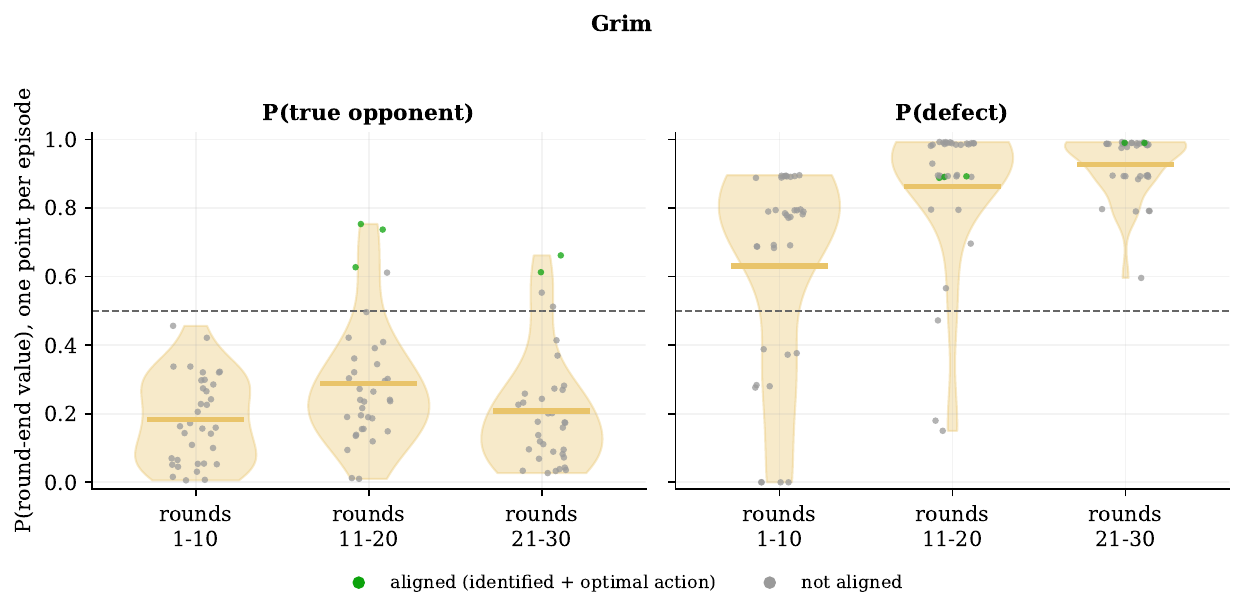}
        \label{fig:round_bin_violin_grim}
    \end{subfigure}
    \vspace{0.5em}
    \begin{subfigure}[b]{\linewidth}
        \centering
        \includegraphics[width=\linewidth]{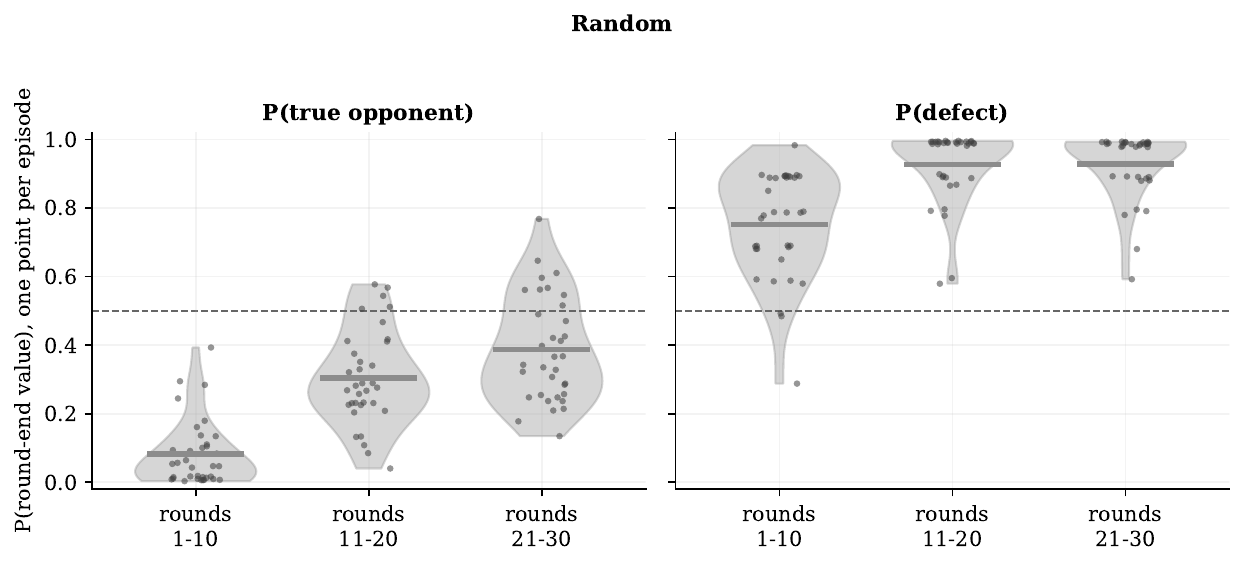}
        \label{fig:round_bin_violin_random}
    \end{subfigure}
    \caption{Round-bin distributions of $P(\text{opp = true})$ (left) and $P(\text{act} = \Def)$ (right) at round-end against the rest of the opponents.}
    \label{fig:round_bin_violin_all}
\end{figure}

Figure~\ref{fig:round_bin_violin_all} extends the per-episode view to the remaining three opponents. \textsc{AllD} appears to be the easiest opponent to identify, with mean $P(\text{opp = true})$ already starting at $0.90$ in rounds 1-10.
Because \textsc{Grim} is behaviorally indistinguishable from \textsc{AllC} until triggered, the model's belief in the true \textsc{Grim} label remains low throughout the game (mean $0.18$--$0.29$ across bins), but $P(\text{act} = \Def)$ steadily climbs from $0.63$ to $0.93$ across rounds. The same pattern applies to \textsc{Random}, where there is by definition no exploitable pattern to lock onto. Here, the mean $P(\text{opp = true})$ remains below chance in all round bins ($0.08 \to 0.30 \to 0.39$), while $P(\text{act} = \Def)$ is already high in the early game ($0.75$), rising further to $0.93$. 
Since the model is uncertain of the opponent's character, it has the proclivity to defect, which is the best strategy for a single-round prisoner's dilemma.

\begin{table*}[t]
\centering
\caption{Word families tracked in the corpus-wide presence-rate analysis, with matched token forms and suppression rate (share of rounds where the family surfaces in the J-lens top-20 but is never emitted in the model's CoT).}
\label{tab:word-families}
\begin{tabular}{lp{0.6\linewidth}r}
\toprule
\textbf{Family} & \textbf{Matched tokens} & \textbf{Suppression\ rate} \\
\midrule
\textsc{giveup} &
\texttt{forever}, \texttt{permanent}, \texttt{permanently} & 68.3\% \\
\textsc{punitive} &
\texttt{punish}, \texttt{punishing}, \texttt{punished}, \texttt{punishment}, \texttt{retaliate}, \texttt{retaliation}, \texttt{retaliating}, \texttt{retaliatory}, \texttt{vengeance} & 70.6\% \\
\textsc{exploited} &
\texttt{exploit}, \texttt{exploiting}, \texttt{exploited}, \texttt{exploitation} & 73.9\% \\
\textsc{reciprocity} &
\texttt{mirror}, \texttt{mirrors}, \texttt{mirroring}, \texttt{mirrored}, \texttt{tit}, \texttt{tat}, \texttt{reciprocal}, \texttt{reciprocate}, \texttt{reciprocity}, \texttt{reciprocated}, \texttt{mutual}, \texttt{forgive}, \texttt{forgiving}, \texttt{forgiven}, \texttt{forgiveness} & 2.4\% \\
\textsc{self-as-victim} &
\texttt{selfish}, \texttt{vulnerable}, \texttt{sucker}, \texttt{naive} & 96.9\% \\
\textsc{hedge} &
\texttt{unlikely}, \texttt{inconsistent}, \texttt{confused}, \texttt{confusing}, \texttt{stuck} & 47.4\% \\
\textsc{risk} &
\texttt{tricky}, \texttt{betray}, \texttt{betrayal}, \texttt{betraying}, \texttt{gamble}, \texttt{gambling}, \texttt{risk}, \texttt{risky} & 46.8\% \\
\bottomrule
\end{tabular}
\end{table*}

\begin{figure*}
    \centering
    \includegraphics[width=\textwidth]{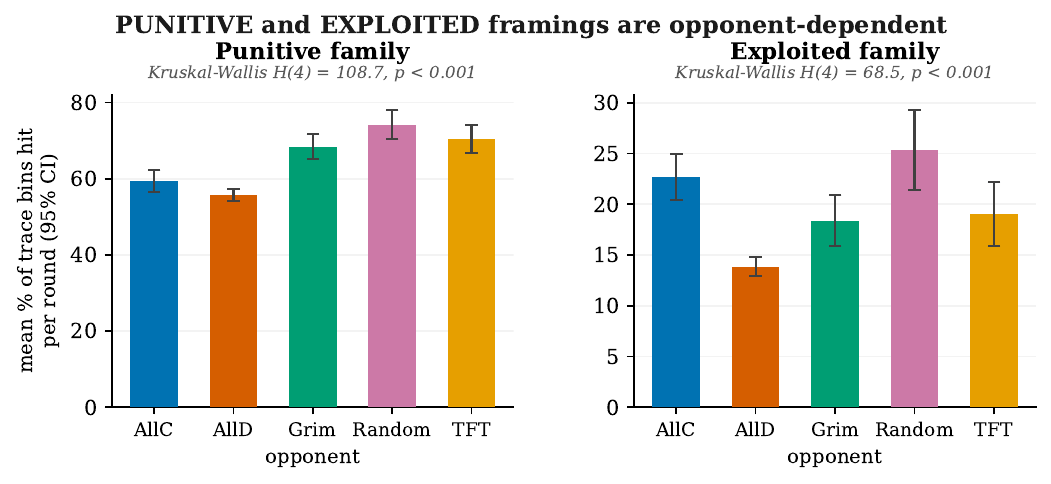}
    \caption{Per-opponent detection rate for the two opponent-dependent concept families from Figure~\ref{fig:silent}, summarized as the mean \% of a round's 40 trace bins in which the family is detected (error bars: $95\%$ CI). Each panel reports a Kruskal-Wallis test of whether this per-round bins-hit count differs by opponent identity (\textsc{punitive}: $H(4){=}108.7$, $p<0.001$; \textsc{exploited}: $H(4){=}68.5$, $p<0.001$), confirming that both families' opponent-dependence is statistically reliable.}
    \label{fig:silent-kw}
\end{figure*}

\end{document}